\documentclass[10pt]{article}

\usepackage[letterpaper]{geometry}
\usepackage{hicss}
\usepackage{times}
\usepackage[none]{hyphenat}
\usepackage{url}
\usepackage{latexsym}
\usepackage{graphicx}
\usepackage{array}
\usepackage{microtype}
\usepackage{booktabs}
\usepackage{tikz}
\usepackage{amsmath}
\usepackage{amssymb}
\usepackage{mathtools}
\usepackage{enumitem}
\usepackage[table]{xcolor}
\usepackage[most]{tcolorbox}
\usepackage[capitalize,noabbrev]{cleveref}
\usetikzlibrary{arrows.meta,positioning,shapes.geometric,fit,calc}

\usepackage[letterpaper]{geometry}
\usepackage{subfigure}
\usepackage{hicss}
\usepackage{times}
\usepackage[none]{hyphenat}
\usepackage{url}
\usepackage{latexsym}
\usepackage{minted}
\usepackage{indentfirst}
\usepackage{graphicx}
\usepackage{array}
\graphicspath{{images/}}
\usepackage[
  style=apa,
]{biblatex}
\graphicspath{{figures/}}

\newtcolorbox[auto counter]{stylebox}[2][]{%
    enhanced,
    float=h,
    breakable,
    colback=blue!3,
    colframe=black!22,
    boxrule=0.5pt,
    arc=3pt,
    left=7pt,
    right=7pt,
    top=9pt,
    bottom=7pt,
    width=\linewidth,
    title={Box~\thetcbcounter: #2},
    fonttitle=\bfseries\footnotesize,
    coltitle=black,
    attach boxed title to top left={
        xshift=7pt,
        yshift=-2.1mm
    },
    boxed title style={
        colback=blue!12,
        colframe=black!25,
        boxrule=0.45pt,
        arc=2pt,
        left=6pt,
        right=6pt,
        top=1.5pt,
        bottom=1.5pt
    },
    #1
}

\title{Building Trustworthy Mental Health Benchmarks on Bluesky: A Validation-Aware Weak-Supervision Framework}

\author{Gaurab Chhetri \\
Department of Computer Science\\
Texas State University \\
San Marcos, Texas, USA\\
{\underline{gaurab@txstate.edu}} \\ \And
Anandi Dutta, Ph.D. \\
Ingram School of Engineering\\
Texas State University \\
San Marcos, Texas, USA\\
{\underline{anandi.dutta@txstate.edu}} \\ \And
Subasish Das, Ph.D. \\
Ingram School of Engineering\\
Texas State University \\
San Marcos, Texas, USA\\
{\underline{subasish@txstate.edu}} \\
}

\date{\today}

\begin{document}
\maketitle

\begin{abstract}
Decentralized social media platforms create new opportunities and challenges for computational mental health research because data access, moderation, labeling, and deployment responsibilities are distributed across multiple technical and governance layers. This paper presents a validation-aware weak-supervision system for constructing and evaluating suicidal ideation (SI) and broader mental health (MH) disclosure benchmarks on Bluesky, a decentralized social media platform built on the AT Protocol. The system integrates public firehose collection, task-specific lexicon filtering, Llama-3-8B-assisted binary annotation, human-adjudicated validation subsets, and transformer-based model benchmarking. Using this pipeline, we construct two task-specific corpora containing 8,346 SI-labeled posts and 9,988 MH-labeled posts. The evaluation shows that model performance depends strongly on both task definition and validation protocol. BERT+LSTM achieves the highest SI stratified cross-validation F1-score, RoBERTa achieves the strongest SI holdout F1-score, and DistilRoBERTa achieves the best MH cross-validation F1-score. Human validation reveals different weak-label failure modes across tasks, with SI labels dominated by false negatives and MH labels dominated by false positives. These findings show that decentralized social media can support reproducible mental health benchmarking, but only when system design, label provenance, validation strategy, and deployment constraints are evaluated together.
\end{abstract}

\subsubsection*{Keywords:}

suicidal ideation detection, mental health NLP, weak-supervision systems, decentralized social media, label validation

\section{Introduction and Background}

Suicide and mental health distress remain major public health concerns, and social media has become an important setting where individuals disclose distress, hopelessness, self-harm thoughts, or related symptoms before seeking formal support \parencite{world2025suicide,coppersmith2018natural,guntuku2017detecting}. Computational mental health research has therefore developed natural language processing (NLP) methods for detecting risk-related signals in user-generated text, especially on centralized platforms such as Reddit, Facebook, and Twitter/X \parencite{de2013predicting,de2014characterizing,shing2018expert}. Although this work shows that online language can support mental health modeling, its assumptions about data access, moderation, governance, and deployment do not directly transfer to decentralized social media. Unlike centralized platforms, decentralized social media separates content hosting, relay infrastructure, moderation, and algorithmic choice across actors, requiring benchmarking to account for data access, label provenance, deployment, and governance together.

\begin{figure}[H]
    \centering
    \includegraphics[width=0.8\linewidth]{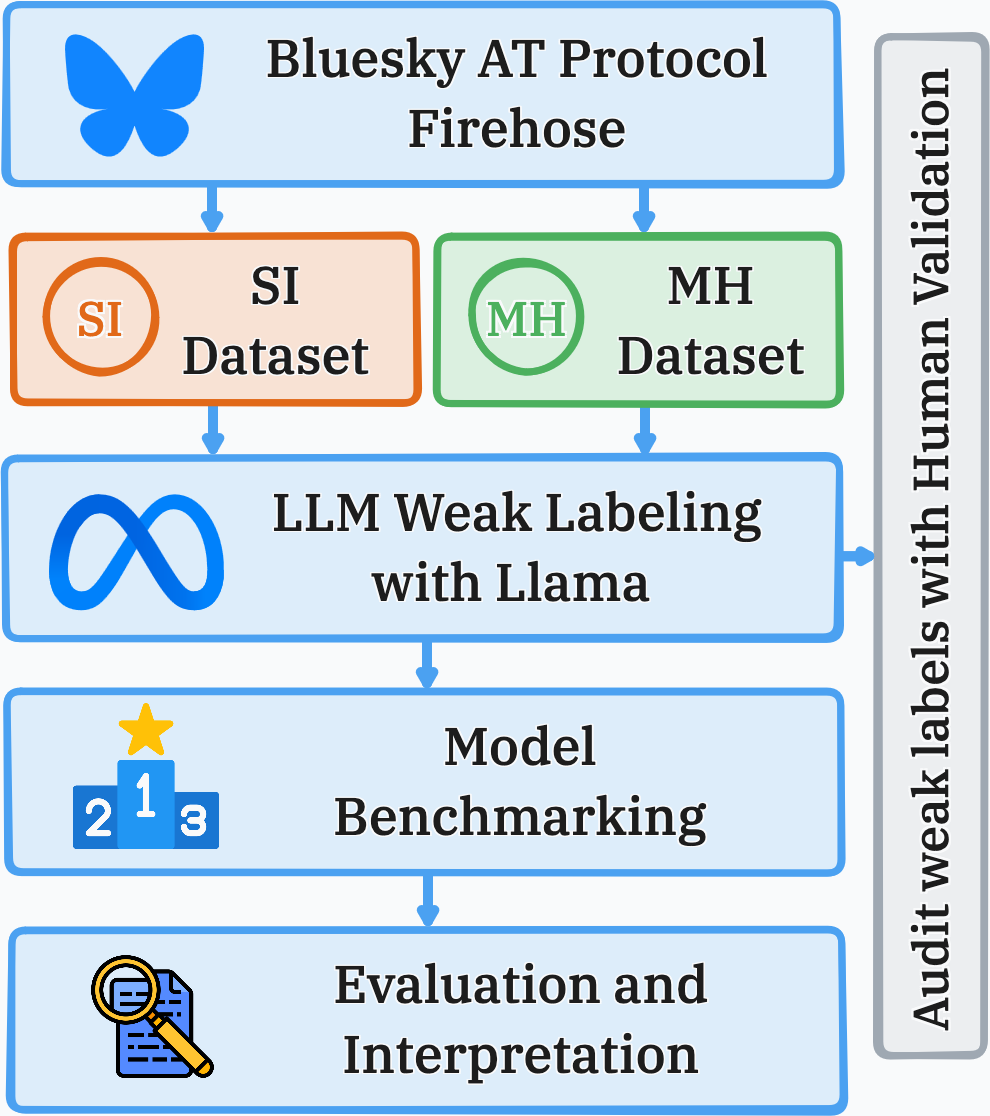}
    \caption{System architecture for validation-aware weak-supervision benchmarking on Bluesky.}
    \label{fig:pipeline}
\end{figure}

Bluesky provides a distinct setting because it is built on the Authenticated Transfer (AT) Protocol, which separates user identity, content hosting, relay infrastructure, and algorithmic choice \parencite{kleppmann2024bluesky}. The AT Protocol firehose enables systematic collection of public posts, but the decentralized architecture also complicates downstream questions about who hosts models, who interprets labels, who acts on predictions, and who is accountable for errors. As shown in \Cref{fig:pipeline}, this study treats mental health disclosure detection as a system-level benchmarking problem that connects data ingestion, candidate retrieval, weak labeling, human validation, model evaluation, and governance-oriented interpretation.

Prior computational mental health studies have used lexical, behavioral, temporal, neural, and transformer-based methods to model depression, distress, and suicidal ideation from social media text \parencite{de2013predicting,tsugawa2015recognizing,coppersmith2018natural,shing2018expert,devlin2019bert,liu2019roberta,sanh2019distilbert}. However, most benchmarks remain centered on centralized platforms such as Reddit and Twitter/X \parencite{gkotsis2017characterisation,turcan2019dreaddit,hasan2024comparative}. Reddit's longer-form posts and community structure differ from short-form decentralized feeds, while X-based studies depend on platform APIs and governance regimes that have changed over time. Bluesky has received growing technical attention because of its open relay infrastructure and portable identity model \parencite{kleppmann2024bluesky}, however mental health disclosure detection remains underexplored in this environment. \textcite{hasan2024comparative} evaluated
a similar SI model set on Reddit. Our contribution therefore lies not in the
model comparison itself, but in a validation-aware benchmark on decentralized
Bluesky data that combines weak supervision, human validation, dual SI/MH
tasks, and deployment-aware analysis.

Mental health detection also requires more than high aggregate model performance. Positive cases may be sparse, disclosures may be ambiguous or figurative. False positives and false negatives carry different risks. Weak supervision from large language models (LLMs) can scale annotation, but it can also introduce systematic label noise. If weak labels are treated as clinical ground truth, model rankings may be inflated or misinterpreted. This is especially important for suicidal ideation (SI) and broader mental health (MH) disclosure detection, where self-directed disclosure must be separated from general discussion, advocacy, humor, third-person narratives, and quoted content. Conceptually, this study is grounded in construct validity and weak-supervision principles, where benchmark reliability depends not only on predictive performance but also on how the target construct is operationalized, labeled, and validated.

This study presents a validation-aware weak-supervision pipeline for benchmarking SI and MH disclosure detection on Bluesky. The paper is a systems evaluation of the full benchmarking pipeline rather than a model leaderboard alone. Box~\ref{box:contributions} summarizes the contributions.

\begin{stylebox}[label={box:contributions}]{Study contributions}
\footnotesize
\begin{itemize}[leftmargin=10pt,itemsep=1pt,topsep=2pt]
\item \textbf{Pipeline:} End-to-end Bluesky collection, weak labeling, validation, and benchmarking.
\item \textbf{Decentralized setting:} Evaluation of access, reproducibility, deployment, and governance on Bluesky.
\item \textbf{Benchmarking:} SI and MH models compared on performance, validation sensitivity, and efficiency.
\item \textbf{Label audit:} Human validation reveals task-specific weak-label errors.
\end{itemize}
\end{stylebox}

The dual-task design is important because SI and MH represent related but conceptually distinct disclosure constructs with different decision boundaries and error consequences. SI detection requires evidence of self-directed intent, desire, planning, or passive death wish, whereas MH disclosure encompasses a broader range of symptoms such as anxiety, depression, panic, insomnia, and hopelessness. Evaluating both tasks under one platform and one weak-supervision framework clarifies which findings are platform-specific, task-specific, or produced by labeling and validation design. The central argument is that decentralized social media benchmarking is a system-level problem. Model performance cannot be interpreted independently of collection architecture, weak-label provenance, human validation, and governance constraints.

\section{Problem Statement}

We formulate SI and MH disclosure detection as supervised binary text classification tasks over public social media posts. Let \( \mathcal{D}^{(k)}=\{(t_i,y_i)\}_{i=1}^{n_k} \) denote the dataset for task \(k\in\{\mathrm{SI},\mathrm{MH}\}\), where \(t_i\) is a post and \(y_i\in\{0,1\}\) is the task-specific label. For SI, \(y_i=1\) denotes self-directed suicidal ideation, including current or recent desire, intent, or planning. For MH, \(y_i=1\) denotes self-directed disclosure of mental health distress or symptoms, including depression, anxiety, hopelessness, panic, or self-harm-related distress.

The objective is to learn a classifier \(f_\theta:\mathcal{T}\rightarrow\{0,1\}\) that maximizes cross-validated F1-score under each task's weak-label regime:
\begin{equation}
f^*=\arg\max_{f_\theta}\mathrm{F1}_{\mathrm{CV}}\left(f_\theta,\mathcal{D}^{(k)}\right),
\end{equation}
where F1 is defined as
\begin{equation}
\mathrm{F1}=2\cdot\frac{\mathrm{Precision}\cdot\mathrm{Recall}}{\mathrm{Precision}+\mathrm{Recall}}
=\frac{2TP}{2TP+FP+FN}.
\end{equation}
F1 is used as the principal metric because both tasks involve imbalanced and safety-sensitive positive classes. Accuracy, precision, recall, Area Under the Curve (AUC), training time, and validation stability are reported where available. Because the two tasks have different error costs, no single metric is sufficient for deployment-oriented interpretation. Precision is relevant when false alarms may stigmatize users or overload human review, while recall is relevant when missed distress could delay support. AUC provides a threshold-independent view of class separability, and training time provides a practical indicator of whether a model can be updated frequently under resource constraints.

Two interpretive assumptions guide the analysis. First, model performance is reported against the label source used for training unless explicitly described as human-adjudicated validation. Thus, cross-validation scores measure consistency with the LLM-generated labeling function, not clinical diagnosis. Second, the unit of analysis is the post rather than the user. The study therefore evaluates whether a single public post contains a task-relevant disclosure, not whether a person is clinically at risk over time.

\section{System Design and Evaluation Methodology}
\label{sec:methodology}

This section describes the design and evaluation of the proposed weak-supervision benchmarking system. As introduced in \Cref{fig:pipeline}, the system links decentralized firehose ingestion, task-specific candidate retrieval, LLM-assisted weak labeling, human validation, model benchmarking, and governance-oriented interpretation. The goal is not only to train classifiers, but also to evaluate how each system layer affects the reliability of SI and broader MH disclosure benchmarks. The requirements in Box~\ref{box:system_requirements} guide the methodological choices described in the following subsections. A companion web page for this work is available at \url{https://ai-in-transportation-lab.github.io/mh-sui-hicss/}.

\begin{stylebox}[label={box:system_requirements}]{System design requirements}
\footnotesize
The pipeline was designed around three requirements:
\begin{itemize}[leftmargin=10pt,itemsep=0pt,topsep=2pt]
\item \textbf{Reproducible ingestion:} collect public posts from a decentralized stream without relying on privileged platform access.
\item \textbf{Scalable annotation with auditability:} use weak supervision for scale while preserving a human-validation layer for construct alignment.
\item \textbf{Deployment-aware evaluation:} report predictive performance together with validation sensitivity, model size, and computational cost.
\end{itemize}
\end{stylebox}

\subsection{Decentralized Data Ingestion and Task Construction}

Both datasets were collected from Bluesky during May--June 2025 using an AT Protocol firehose pipeline adapted from the CognitiveSky framework \parencite{chhetri2025cognitivesky}. Unlike conventional social media APIs that rely on centralized rate limits and privileged access, the AT Protocol provides a public stream of posts through relay infrastructure. This setting supports reproducible data collection, but it also requires careful interpretation because public availability does not imply clinical validity or consent for individual-level intervention. The SI corpus was collected over a 14-day period using a suicide-related lexicon derived from computational mental health literature and cross-checked against established suicidality datasets \parencite{coppersmith2018natural,guntuku2017detecting,shing2018expert,turcan2019dreaddit}. The resulting dataset contains 8,346 posts from 6,648 unique users, including 2,180 SI-positive posts (26.1\%) and 6,166 SI-negative posts (73.9\%). The MH corpus was collected over a two-month period using a broader mental-health lexicon and contains 9,988 labeled posts, including 4,793 MH-positive posts and 5,195 MH-negative posts. Because the retrieval criteria and collection durations differed, substantial post-level overlap is unlikely;
however, cross-corpus overlap was not explicitly quantified.

Keyword-triggered collection was used to enrich the stream for potentially relevant content rather than to estimate platform-level prevalence. For SI, the lexicon focused on both explicit self-harm and death-wish expressions and indirect indicators such as hopelessness and perceived burdensomeness. The lexicon for MH included broader symptom and distress terms related to depression, anxiety, panic, insomnia, and help-seeking. This design increases the density of relevant cases, but it also introduces selection bias. Posts expressing distress without matching the lexicon may be absent, while posts discussing mental health in news, advocacy, or humorous contexts may be overrepresented. Only publicly available Bluesky posts were used, and the benchmark is intended for aggregate research and model evaluation rather than clinical diagnosis or automated individual-level intervention. Given the sensitivity of SI and MH disclosures, results are interpreted at the post level and with explicit attention to privacy, label uncertainty, and responsible use of public social media data.

\subsection{Weak-Supervision Labeling Layer}

Labels were generated at scale using Meta's Llama-3-8B through a local Ollama-based annotation pipeline \parencite{dubey2024llama}. Prompts instructed the model to return only \texttt{1} for positive cases or \texttt{0} for negative cases. This deterministic binary-output design reduced response ambiguity and simplified parsing at scale. Each post was labeled independently without user history, reply context, profile metadata, or temporal sequences. Throughout the paper, model scores are interpreted as performance against the weak-labeling function unless explicitly evaluated against human-adjudicated labels. Weak supervision enables scalable benchmarking, but it also means that label provenance must be evaluated alongside model performance. Consequently, the pipeline includes a separate human-validation layer rather than treating LLM labels as ground truth.

\subsection{Human Validation Layer}

Human validation was conducted on stratified random subsets of 276 SI posts and 250 MH posts, with both positive and negative weak-label classes represented. All validation items were independently labeled by two annotators with backgrounds in computing and applied AI; neither annotator had formal clinical training. Both annotators applied the same binary guidelines, requiring explicit self-directed disclosure for positive labels and excluding news, advocacy, third-person narratives, hypothetical statements, jokes, and ambiguous mentions. Thus, every validation item was doubly annotated. Human-human inter-annotator reliability was measured using Cohen's $\kappa$ from the two annotators' independent labels before adjudication. Disagreements were subsequently reviewed jointly against the task definitions and boundary rules, and a consensus label was retained as the adjudicated human reference. The LLM-generated weak labels were then evaluated against these adjudicated labels using observed agreement, Cohen's $\kappa$, positive-class precision, recall, F1-score, and macro-F1. Cohen's $\kappa$ was calculated as
\begin{equation}
\label{eq:kappa}
\kappa=\frac{p_o-p_e}{1-p_e},
\end{equation}
where $p_o$ is the observed proportion of agreement and
$p_e$ is the agreement expected by chance from the marginal label
distributions. \Cref{tab:human_validation} reports both human-human reliability
and LLM--human agreement measures.

\begin{table}[h]
\caption{Human validation of LLM-assisted weak labels.}
\label{tab:human_validation}
\centering
\footnotesize
\setlength{\tabcolsep}{3pt}
\begin{tabular}{lcc}
\toprule
\rowcolor{blue!6}
\textbf{Measure} & \textbf{SI} & \textbf{MH} \\
\midrule
Validation sample & 276 & 250 \\
Human-human Cohen's ($\kappa$) & 0.762 & 0.595 \\
LLM-human observed agreement & 88.8\% & 55.6\% \\
LLM-human Cohen's ($\kappa$) & 0.636 & 0.222 \\
LLM-human precision & 0.947 & 0.276 \\
LLM-human recall & 0.554 & 0.977 \\
Positive-class F1 & 0.699 & 0.431 \\
Macro-F1 & 0.815 & 0.533 \\
Dominant noise mode & false negative & false positive \\
\bottomrule
\end{tabular}
\end{table}

The validation results in \Cref{tab:human_validation} show substantially
different weak-label behavior across the two tasks. For SI, the LLM correctly
classified 209 non-SI posts and 36 SI posts, while missing 29 SI posts and
falsely labeling 2 non-SI posts as SI. This corresponds to an observed
LLM--human agreement of 88.8\%, a chance-corrected Cohen's $\kappa$ of 0.636,
precision of 0.947, and recall of 0.554. The dominant SI error mode is
therefore false negatives under the human definition of self-directed suicidal
ideation. For MH, the LLM correctly identified 97 non-MH posts and 42 MH posts, while
producing 110 false positives and 1 false negative. Observed LLM--human
agreement was 55.6\%, but chance-corrected agreement was substantially lower
($\kappa=0.222$). The weak labeler achieved high recall (0.977) but low
precision (0.276), indicating systematic over-inclusion of general mental
health discourse, advocacy, metaphor, and third-person discussion. The
difference between observed and chance-corrected agreement is particularly
important for MH because the LLM labeled 60.8\% of the validation sample as
positive, whereas only 17.2\% was positive after human adjudication.

\begin{table*}[t]
\caption{Annotation boundary rules used for human validation.}
\label{tab:annotation_rules}
\centering
\footnotesize
\setlength{\tabcolsep}{4pt}
\begin{tabular}{p{0.09\linewidth}p{0.40\linewidth}p{0.40\linewidth}}
\toprule
\rowcolor{blue!6}
\textbf{Category} & \textbf{Positive examples in scope} & \textbf{Excluded cases} \\
\midrule
SI & First-person desire to die, passive death wish, current self-harm ideation, planning language, or recent self-directed suicidal intent. & News about suicide, advocacy posts, condolences, quotations, jokes, hypotheticals, and third-person discussion without self-directed ideation. \\
MH & First-person disclosure of depression, anxiety, panic, hopelessness, distress-related insomnia, help-seeking, or self-harm-related distress. & General awareness posts, clinical information sharing, political commentary, metaphors, sarcasm, and statements about another person's mental health. \\
Ambiguous content & Labeled positive only when self-directed disclosure was explicit enough to support the task definition. & Labeled negative when context was insufficient, quoted, purely rhetorical, or not attributable to the author. \\
\bottomrule
\end{tabular}
\end{table*}

\subsection{Training and Evaluation Protocols}

The benchmarking stage evaluates whether different model families reproduce the task-specific weak-labeling functions under practical deployment constraints. Rather than treating model ranking as the sole objective, the evaluation compares high-capacity encoders, compact encoders, and hybrid sequence models to test whether conclusions remain stable across architectures and computational profiles. The SI benchmark includes BERT, RoBERTa, DistilBERT, ALBERT, ELECTRA, XLNet, MobileBERT, and two BERT+LSTM variants. The MH benchmark focuses on deployable compact models, including DistilRoBERTa, DistilBERT, MiniLM, ELECTRA-small, and TinyBERT. All transformer models were implemented with HuggingFace Transformers \parencite{wolf2020transformers}. All transformer models were fine-tuned using AdamW with learning rate ($2\times10^{-5}$). The SI transformer experiments used 3 epochs, batch size 16 for training, and batch size 64 for evaluation. The SI hybrid models used Adam with learning rate ($1\times10^{-3}$), weight decay ($1\times10^{-4}$), batch size 16, and early stopping with patience 3 over at most 5 epochs. The MH experiments used stratified 5-fold cross-validation, up to 5 epochs, early stopping with patience 2, maximum sequence length 512, dynamic padding, batch size 8 for training, and batch size 16 for evaluation. All experiments were conducted on macOS hardware with an Apple M4 Pro processor and 24 GB unified memory.

\Cref{tab:evaluation_design} summarizes the task-specific evaluation design. SI results include both a 70/30 stratified holdout split and cross-validation because safety-sensitive model selection can be sensitive to data partitioning. The holdout split provides a deployment-like estimate on unseen data, while cross-validation provides a lower-variance comparison across architectures. MH results use stratified 5-fold cross-validation as the primary protocol because the goal is to compare compact models under a consistent evaluation setting. Across both tasks, the evaluation reports F1-score as the primary metric, along with accuracy, precision, recall, AUC where available, and training time as a practical proxy for computational cost.

\begin{table}[h]
\caption{Task-specific evaluation design.}
\label{tab:evaluation_design}
\centering
\footnotesize
\setlength{\tabcolsep}{3pt}
\begin{tabular}{lcc}
\toprule
\rowcolor{blue!6}
\textbf{Design element} & \textbf{SI} & \textbf{MH} \\
\midrule
Primary validation & Holdout + CV & 5-fold CV \\
Optimization metric & F1 & F1 \\
Model emphasis & broad architecture set & compact encoders \\
Transformer epochs & 3 & up to 5 \\
Early stopping & hybrid models & all models \\
Train batch size & 16 & 8 \\
Evaluation batch size & 64 / 32 & 16 \\
Efficiency measure & train time & per-fold time \\
\bottomrule
\end{tabular}
\end{table}

\section{System Evaluation Results}
\label{sec:results}

This section evaluates the proposed benchmarking system across four dimensions: corpus construction, model performance, validation sensitivity, and operational efficiency. The goal is not only to identify the highest-scoring classifier, but also to determine how each stage of the pipeline affects the reliability and interpretation of the final benchmark.

\subsection{Dataset Characteristics}

\Cref{tab:dataset_summary} summarizes the two Bluesky corpora generated by the collection and weak-labeling pipeline. The SI corpus is smaller and more imbalanced, reflecting a narrower construct centered on self-directed suicidal ideation. The MH corpus is larger and more balanced because the broader lexicon captures a wider range of distress-related disclosures. The dataset patterns in \Cref{tab:dataset_summary} show that task construction impact downstream interpretation before modeling begins. The SI corpus contains an average of 1.3 posts per unique user, suggesting that the benchmark is not dominated by a small number of highly active accounts. However, the post-level design also means that the system does not use longitudinal information, reply context, or user history. The benchmark is therefore appropriate for evaluating text-level disclosure detection, but not for estimating person-level clinical risk. 

The different positive rates also reflect the distinction between the two constructs. SI has a lower positive rate despite a suicide-focused lexicon because many posts mentioning suicide refer to public events, fictional content, jokes, news, or other people. MH has a higher positive rate because broad distress language is more common and because the positive class includes a wider set of symptoms and disclosures. This difference explains why MH models can obtain high cross-validation scores even when human agreement is weaker. The weak-labeling function captures a broad discourse category that is easier for models to reproduce than for human annotators to accept as strict self-disclosure.

\begin{table}[h]
\caption{Dataset summary for the two Bluesky benchmarks.}
\label{tab:dataset_summary}
\centering
\footnotesize
\setlength{\tabcolsep}{4pt}
\begin{tabular}{lcc}
\toprule
\rowcolor{blue!6}
\textbf{Property} & \textbf{SI} & \textbf{MH} \\
\midrule
Collection window & 14 days & 2 months \\
Labeled posts & 8,346 & 9,988 \\
Positive posts & 2,180 & 4,793 \\
Negative posts & 6,166 & 5,195 \\
Positive rate & 26.1\% & 48.0\% \\
Unique users & 6,648 & not reported \\
\bottomrule
\end{tabular}
\end{table}

\subsection{Suicidal Ideation Benchmark}

\Cref{tab:si_holdout} reports holdout validation performance for the nine SI models. RoBERTa-base achieves the strongest holdout F1-score (0.792), followed by ELECTRA-base (0.784), ALBERT-base-v2 (0.778), BERT-base (0.774), and DistilBERT-base (0.773). DistilBERT provides the strongest efficiency trade-off among the top transformer models, reaching near-RoBERTa performance with roughly half the training time. The holdout results in \Cref{tab:si_holdout} reveal important precision-recall differences. BERT+LSTM+Attention produces the highest recall (0.870), but its lower precision reduces overall F1. This trade-off matters in sensitive settings because high recall may be desirable only when human review is available and false-positive burden is manageable. RoBERTa and ELECTRA provide more balanced holdout behavior, while DistilBERT offers the most favorable speed-performance profile among the high-performing transformer encoders. \Cref{tab:si_cv} reports SI cross-validation performance. In contrast to the holdout setting, BERT+LSTM achieves the highest F1-score across all cross-validation variants, reaching 0.916 under stratified 5-fold cross-validation. This contrast indicates that validation protocol changes the apparent best model.

\begin{table*}[h]
\caption{SI holdout validation performance.}
\label{tab:si_holdout}
\centering
\footnotesize
\setlength{\tabcolsep}{4pt}
\begin{tabular}{lccccc}
\toprule
\rowcolor{blue!6}
\textbf{Model} & \textbf{Accuracy} & \textbf{Precision} & \textbf{Recall} & \textbf{F1} & \textbf{Train time (s)} \\
\midrule
RoBERTa-base & \textbf{0.892} & 0.799 & 0.784 & \textbf{0.792} & 841 \\
ELECTRA-base & 0.891 & 0.813 & 0.757 & 0.784 & 834 \\
ALBERT-base-v2 & 0.886 & 0.789 & 0.768 & 0.778 & 908 \\
BERT-base & 0.883 & 0.785 & 0.763 & 0.774 & 835 \\
DistilBERT-base & 0.883 & 0.780 & 0.766 & 0.773 & \textbf{423} \\
XLNet-base & 0.884 & 0.797 & 0.745 & 0.770 & 1359 \\
MobileBERT & 0.862 & 0.775 & 0.665 & 0.716 & 563 \\
BERT+LSTM & 0.892 & \textbf{0.873} & 0.687 & 0.769 & 4524 \\
BERT+LSTM+Attention & 0.834 & 0.633 & \textbf{0.870} & 0.733 & 4512 \\
\bottomrule
\end{tabular}
\end{table*}

\begin{table}[h]
\caption{SI cross-validation performance.}
\label{tab:si_cv}
\centering
\footnotesize
\setlength{\tabcolsep}{6pt}
\begin{tabular}{lcccc}
\toprule
\rowcolor{blue!6}
\textbf{Model} &
\multicolumn{2}{c}{\textbf{Stratified 5-fold}} &
\multicolumn{2}{c}{\textbf{Repeated 5-fold}} \\
\cmidrule(lr){2-3}\cmidrule(lr){4-5}
&
\textbf{Mean F1} & \textbf{SD} &
\textbf{Mean F1} & \textbf{SD} \\
\midrule
BERT+LSTM
& \textbf{0.9162} & \textbf{0.007}
& \textbf{0.9148} & 0.009 \\

RoBERTa-base & 0.9017 & 0.012 & 0.9024 & 0.010 \\
BERT-base    & 0.8986 & 0.006 & 0.8993 & 0.010 \\

XLNet-base
& 0.8904 & 0.008
& 0.8896 & 0.014 \\

ELECTRA-base
& 0.8863 & 0.006
& 0.8860 & 0.016 \\

ALBERT-base-v2
& 0.8833 & 0.011
& 0.8828 & 0.014 \\

BERT+LSTM+Attn.
& 0.8791 & 0.010
& 0.8783 & \textbf{0.008} \\

DistilBERT-base
& 0.8712 & 0.006
& 0.8705 & 0.015 \\

MobileBERT
& 0.7844 & 0.013
& 0.7840 & 0.014 \\
\bottomrule
\end{tabular}
\end{table}

\begin{figure}[h]
\centering
\includegraphics[width=0.8\linewidth]{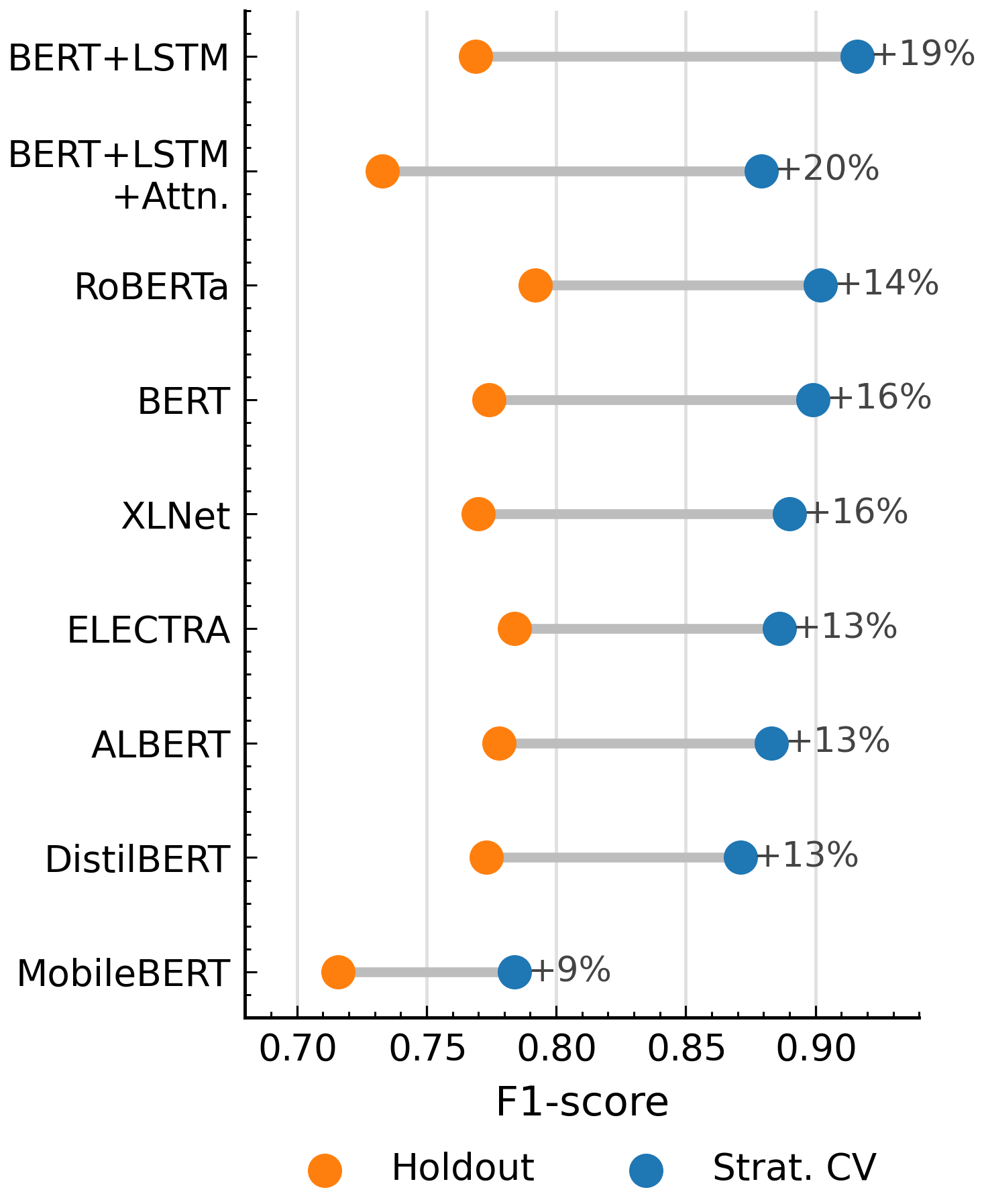}
\caption{SI holdout F1 versus stratified cross-validation F1.}
\label{fig:si_holdout_cv}
\end{figure}

\begin{figure}[h]
\centering
\includegraphics[width=0.8\linewidth]{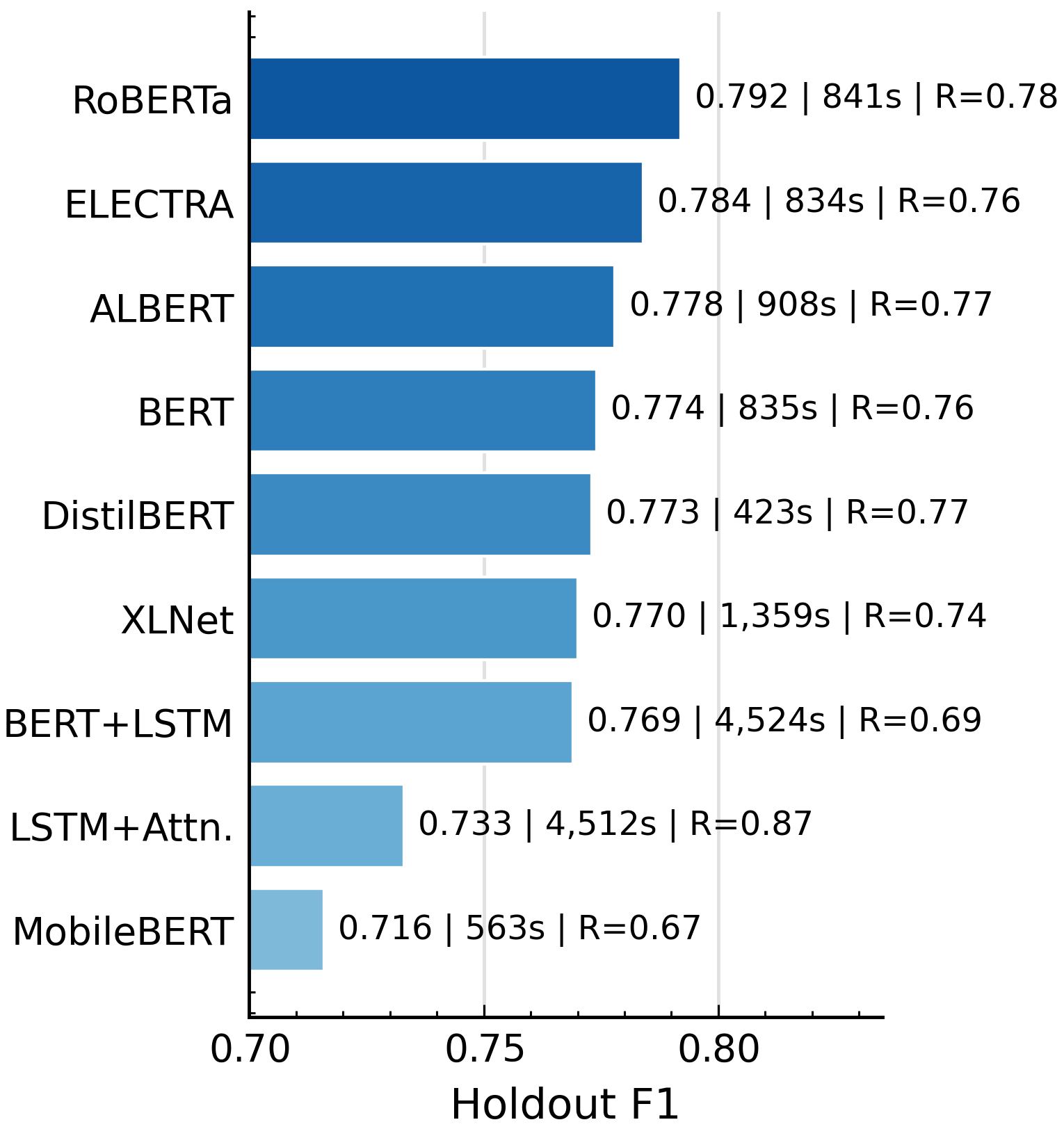}
\caption{SI efficiency summary. Bars show holdout F1; right-side labels report F1, training time, and recall for every model.}
\label{fig:si_efficiency}
\end{figure}

The validation-protocol effect is visualized in \Cref{fig:si_holdout_cv}. Every SI model improves under stratified cross-validation relative to holdout evaluation, but the gain is not uniform. Hybrid models benefit the most, suggesting that they take advantage of the larger effective training exposure across folds. MobileBERT gains the least, indicating that compact architectures may have less capacity to exploit additional partition diversity. \Cref{fig:si_efficiency} summarizes the SI efficiency profile by placing holdout F1, training time, and recall in the same view. The figure reinforces the practical distinction between maximum performance and usable performance: RoBERTa leads holdout F1, BERT+LSTM+Attention leads recall, and DistilBERT provides the strongest computational value among the top transformer models. The system does not identify a single universally best model. Instead, it exposes how validation protocol, recall requirements, and computational cost change the preferred model.

\subsection{Mental Health Disclosure Benchmark}

\Cref{tab:mh_results} reports MH cross-validation performance against the
LLM-generated weak labels for five compact transformer models. DistilRoBERTa
achieves the highest weak-label F1-score (0.905), with accuracy 0.909 and
AUC 0.931. These scores quantify reproduction of the weak-labeling
function rather than performance against the human-adjudicated MH construct;
the weak-labeling function itself achieved only 0.431 positive-class F1
against human adjudication (\Cref{tab:human_validation}). DistilBERT and MiniLM follow closely, indicating that compact encoders preserve most of the classification performance under the MH weak-label regime. The small spread among the top three MH models in \Cref{tab:mh_results} is more important than the rank order alone. DistilRoBERTa leads DistilBERT by only 0.002 F1 and MiniLM by only 0.004 F1, while requiring substantially more training time than MiniLM. This suggests that deployment requirements should guide model choice. If the goal is maximum average discrimination, DistilRoBERTa is preferred. For rapid iteration on limited hardware, MiniLM provides nearly equivalent F1 at much lower cost. TinyBERT is the fastest model, but its lower F1 and AUC suggest that aggressive compression may reduce robustness to heterogeneous MH disclosure language.

\begin{figure}[h]
\centering
\includegraphics[width=0.8\linewidth]{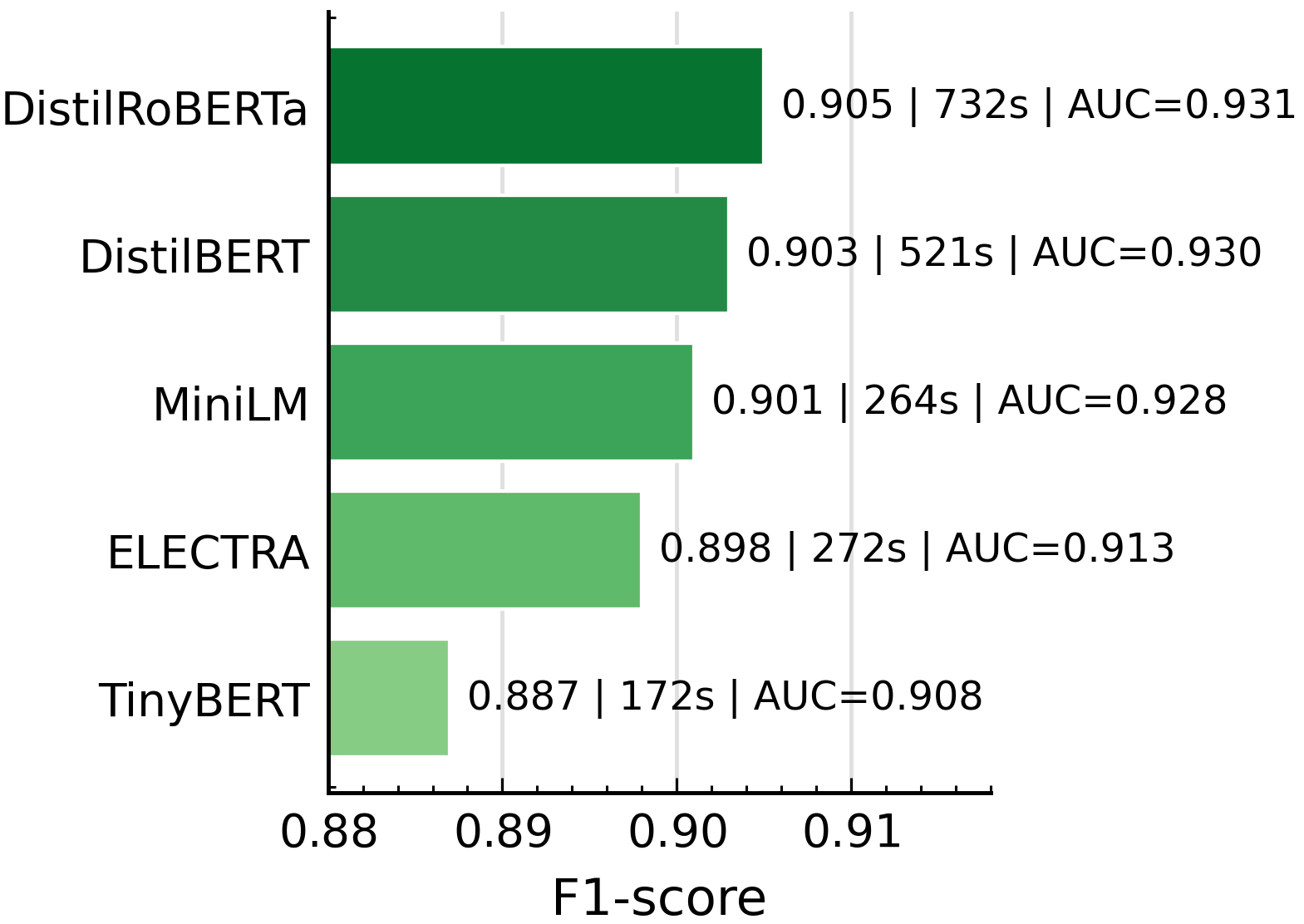}
\caption{MH efficiency summary. Bars show F1-score; right-side labels report F1, training time, and AUC for every model.}
\label{fig:mh_efficiency}
\end{figure}

\Cref{fig:mh_efficiency} shows the corresponding efficiency profile for MH models. The figure makes the speed-performance trade-off visible. DistilRoBERTa is strongest by F1 and AUC, MiniLM offers the clearest speed-performance balance, and TinyBERT minimizes training time. AUC values above 0.90 for all MH models indicate strong separability under the weak-label regime. However, these scores must be read together with the human validation results reported earlier in \Cref{tab:human_validation}. Because the LLM labeler over-predicts MH disclosure relative to human adjudication. High AUC means that models separate LLM-positive from LLM-negative examples well. It does not mean that the models have solved strict human-grounded disclosure detection.

\begin{table}[h]
\caption{MH cross-validation performance (5-fold mean).}
\label{tab:mh_results}
\centering
\footnotesize
\setlength{\tabcolsep}{3pt}
\begin{tabular}{lccccc}
\toprule
\rowcolor{blue!6}
\textbf{Model} & \textbf{F1} & \textbf{Acc.} & \textbf{AUC} & \textbf{Prec.} & \textbf{Time (s)} \\
\midrule
DistilRoBERTa & \textbf{0.905} & \textbf{0.909} & \textbf{0.931} & \textbf{0.903} & 731.78 \\
DistilBERT & 0.903 & 0.907 & 0.930 & 0.901 & 521.05 \\
MiniLM & 0.901 & 0.904 & 0.928 & 0.891 & 263.84 \\
ELECTRA & 0.898 & 0.898 & 0.913 & 0.865 & 272.00 \\
TinyBERT & 0.887 & 0.890 & 0.908 & 0.877 & \textbf{171.51} \\
\bottomrule
\end{tabular}
\end{table}

\subsection{Cross-Task Synthesis}

The SI and MH tasks show different relationships between label quality and model performance. SI labels have stronger human agreement but lower positive prevalence, making F1 sensitive to missed positive cases. MH labels are broader and more balanced, but human validation reveals substantial false-positive noise from advocacy, metaphor, third-person narration, and general mental health discussion. \Cref{tab:cross_task} summarizes these cross-task differences. The comparison in \Cref{tab:cross_task} highlights a central systems finding. SI and MH cannot be treated as interchangeable detection tasks. SI performance more directly targets a narrow self-disclosure construct, but it remains vulnerable to subtle false negatives. MH performance captures a broader weak-labeling function, but that function is more likely to over-include general discourse. As a result, the SI benchmark is more sensitive to missed-risk interpretation, while the MH benchmark is more sensitive to over-classification. The benchmark is most informative when the two tasks are interpreted comparatively rather than collapsed into a single mental health detection problem.

\begin{table*}[h]
\caption{Interpretive comparison of SI and MH benchmark outcomes.}
\label{tab:cross_task}
\centering
\footnotesize
\setlength{\tabcolsep}{4pt}
\begin{tabular}{p{0.18\linewidth}p{0.36\linewidth}p{0.36\linewidth}}
\toprule
\rowcolor{blue!6}
\textbf{Dimension} & \textbf{SI benchmark} & \textbf{MH benchmark} \\
\midrule
Construct boundary & Narrower; requires self-directed suicidal ideation or passive desire to die. & Broader; includes self-directed mental health symptoms and distress. \\
Dominant validation risk & False negatives for subtle, indirect, or context-dependent SI. & False positives for awareness, advocacy, metaphor, humor, and third-person discussion. \\
Best performance signal & BERT+LSTM under stratified CV; RoBERTa under holdout. & DistilRoBERTa under stratified 5-fold CV. \\
Efficiency signal & DistilBERT gives strong holdout performance at lowest transformer training time. & TinyBERT is fastest, while MiniLM offers a strong speed-performance balance. \\
Deployment implication & Requires conservative interpretation and high-sensitivity review workflows if used for screening research. & Requires calibration to avoid over-classifying general discourse as self-disclosure. \\
\bottomrule
\end{tabular}
\end{table*}

\subsection{Error Type Analysis}

Human validation indicates that weak-label errors are structured rather than random. SI false negatives often involve short posts, indirect death-wish language, quoted or context-dependent statements, and ambiguous expressions of despair that human annotators treat as self-directed but the LLM does not. These cases are difficult because the relevant signal may be pragmatic rather than lexical. MH false positives follow a different pattern. Many contain mental-health vocabulary but do not constitute self-disclosure. Examples include awareness campaigns, discussion of public figures, support messages directed at others, and metaphorical uses of terms such as `depressing' or `panic.' 

The main weak-label error sources were advocacy or awareness posts, third-person narratives, metaphorical use of clinical vocabulary, sarcasm or humor, indirect ideation, and missing conversational context. These errors have different implications across tasks. MH false positives often arise from broad topical discussion, while SI false negatives often arise when self-directed intent is indirect or context-dependent. The error sources motivate a two-stage interpretation of the benchmark. The first question is whether standard models can learn the scalable weak-labeling function. The reported performance results show that they can. The second question is whether that function aligns with the intended human construct. Human validation shows partial alignment for SI and weaker alignment for MH. Therefore, high predictive performance should be reported together with evidence about label provenance and construct validity.

\subsection{Deployment and Governance Implications}

The decentralized setting changes how model outputs might be used. In Bluesky-style architectures, content hosting, relays, clients, labelers, and user-selected algorithms may be operated by different actors. A mental health classifier could therefore function as a research instrument, client-side filter, community labeler, or aggregate monitoring tool, each with different governance requirements.

For SI, the most defensible near-term use is aggregate research and tool evaluation rather than automated user intervention. Any individual-level use would require human review, crisis-response protocols, opt-in governance where appropriate, and clear accountability for false positives and false negatives. For MH disclosure detection, automated intervention is even less appropriate because many positive weak labels correspond to public discourse rather than genuine personal distress.

The efficiency results in \Cref{fig:si_efficiency} and \Cref{fig:mh_efficiency} are relevant to governance because model size affects who can run, audit, and update the system. DistilBERT, MiniLM, and TinyBERT lower the barrier for independent researchers or community operators. However, efficiency should not be confused with readiness for intervention. Compact models can reproduce weak labels quickly, but sensitive deployment still requires human-grounded calibration, transparent oversight, and task-specific validation.

\section{Discussion}

This study shows that mental health disclosure benchmarking on decentralized social media is a system-level problem rather than a classifier-selection problem alone. Benchmark reliability depends on the full pipeline in \Cref{fig:pipeline}, including AT Protocol access, lexicon-based retrieval, LLM-assisted weak labeling, human validation, model evaluation, and governance-oriented interpretation. Bluesky's public infrastructure improves reproducibility, but does not ensure construct validity. Human validation in \Cref{tab:human_validation} shows that label quality differs substantially by task. SI labels showed stronger agreement, although false negatives remained important. MH weak labels showed only 55.6\% observed agreement with the human-adjudicated reference and Cohen's $\kappa=0.222$, with very high recall (0.977) but low precision (0.276). This indicates systematic over-inclusion of mental-health-related discourse and means the MH benchmark should be interpreted primarily as consistency with the weak-labeling function rather than direct performance against a strongly human-aligned disclosure construct.

Weak supervision must therefore be treated as a visible dataset property. Model scores primarily measure agreement with the LLM-assisted labeling function unless evaluated against human-adjudicated labels, so high benchmark performance should not be interpreted as clinical validity. Recurring errors such as advocacy, third-person discussion, metaphor, sarcasm, indirect ideation, and missing context further show that label noise is structured rather than random. Model rankings also depend on validation protocol and operational constraints. For SI, RoBERTa has the highest holdout performance in \Cref{tab:si_holdout}, whereas BERT+LSTM leads cross-validation in \Cref{tab:si_cv}. For MH, DistilRoBERTa has the highest F1 in \Cref{tab:mh_results}, but DistilBERT and MiniLM are close enough that efficiency may matter more than small performance differences. Overall, the SI benchmark has stronger human-validation support, while the MH benchmark remains more preliminary. These findings reinforce the central contribution of the framework: predictive performance should be interpreted together with task definition, label provenance, human validation, and deployment context.

\section{Limitations and Future Work}

The first limitation is keyword-triggered retrieval. Although lexicon-based collection increases the density of relevant SI and MH content, it may miss implicit, coded, or context-dependent disclosures, particularly SI expressed indirectly or across reply chains. Future work should evaluate semantic retrieval, classifier-in-the-loop sampling, and active learning to improve coverage. A second limitation concerns label validity because human validation covered only stratified subsets and could not capture all weak-label errors. The limited chance-corrected LLM--human agreement for MH
($\kappa=0.222$) further indicates that MH model scores should not be
interpreted as estimates of performance against a human-grounded disclosure
construct. Future benchmarks should include larger human-grounded test sets, diverse annotators, and clearer distinctions among personal disclosure, general discussion, support-seeking, advocacy, and third-person narration. Future work should systematically compare alternative weak-labeling strategies, LLM labelers, prompt formulations, and retrieval methods to determine whether the observed findings are robust to different framework implementations.

The third limitation is that the benchmark is post-level rather than longitudinal and therefore does not model user trajectories, repeated disclosures, escalation, social responses, or temporal language changes. Although longitudinal modeling may improve risk-sensitive analysis, it also raises stronger privacy, consent, and governance concerns; future work should assess temporal features without enabling intrusive surveillance or individual-level targeting. Fourth, the evaluation emphasizes predictive performance and efficiency but lacks interpretability and component-level ablations. Token attribution, counterfactual editing, error analysis, and attention-based diagnostics could reveal whether models rely on meaningful cues or superficial lexical triggers, while ablations of retrieval, weak-label prompts, validation sampling, and temporal splits could clarify failure propagation. Finally, the results reflect Bluesky during the collection period and may change with platform language, demographics, affordances, and moderation practices. Replication across collection windows and other AT Protocol services is therefore needed before treating the findings as stable estimates of decentralized social media performance.

\section{Conclusions}

This study presents and evaluates a validation-aware weak-supervision pipeline for SI and MH disclosure detection on Bluesky. By combining AT Protocol firehose collection, task-specific retrieval, Llama-3-8B-assisted labeling, human validation, and transformer-based benchmarking, the paper shows that decentralized social media can support reproducible mental health disclosure benchmarks. The results also show why such benchmarks must be interpreted carefully. BERT+LSTM leads SI cross-validation, RoBERTa leads SI holdout evaluation, and DistilRoBERTa leads MH detection. However, these rankings depend on task definition, label provenance, validation protocol, and computational constraints. These rankings should be interpreted descriptively rather than as evidence of statistically significant differences between models. The central conclusion is that responsible benchmarking for sensitive mental health tasks requires evaluating the full system around the model, not only the model itself.

\printbibliography

\end{document}